\pdfoutput=1

\documentclass[11pt]{article}

\usepackage[final]{acl}

\usepackage{latexsym}
\usepackage{subfigure}
\usepackage[T1]{fontenc}
\usepackage{booktabs}
\usepackage[utf8]{inputenc}

\usepackage{microtype}

\usepackage{inconsolata}

\usepackage{graphicx}
\usepackage{times}
\usepackage{epsfig}
\usepackage{amsmath}
\usepackage{amssymb}

\usepackage{multirow}
\usepackage{array} 
\newcolumntype{C}[1]{>{\centering}p{#1}} 

\usepackage{colortbl}
\title{Advancing Sequential Numerical Prediction in Autoregressive Models}

\author{%
  Xiang Fei\thanks{Equal Contribution \newline \hspace*{0.40cm}\dag Corresponding Author}$^{1}$ \ \ Jinghui Lu$^{* 1}$ \ \ Qi Sun$^{* 2}$ \ \ Hao Feng$^{\dag 1}$ \ \ Yanjie Wang$^{1}$ \\
  \textbf{Wei Shi$^{1}$ \ \ An-lan Wang$^{1}$ \ \ Jingqun Tang$^{1}$  \ \ Can Huang$^{\dag 1}$}
  \\
  $^1$ByteDance Inc. \ \ $^2$City University of Hong Kong\\
  \texttt{\{feixiang.77, lujinghui, fenghao.2019\}@bytedance.com}\\
  \texttt{\{wangyanjie.prince, shiwei.11, wanganlan\}@bytedance.com}\\
  \texttt{\{tangjingqun, can.huang\}@bytedance.com} \\
  \texttt{qisun.new@gmail.com}
}

\begin{document}
\maketitle
\begin{abstract}

Autoregressive models have become the de facto choice for sequence generation tasks, but standard approaches treat digits as independent tokens and apply cross-entropy loss, overlooking the coherent structure of numerical sequences. This paper introduces \textit{\textbf{N}umerical \textbf{T}oken \textbf{I}ntegrity \textbf{Loss} (NTIL)} to address this gap. NTIL operates at two levels: (1) token-level, where it extends the Earth Mover's Distance (EMD) to preserve ordinal relationships between numerical values, and (2) sequence-level, where it penalizes the overall discrepancy between the predicted and actual sequences. This dual approach improves numerical prediction and integrates effectively with LLMs/MLLMs. Extensive experiments show significant performance improvements with NTIL. All resources are available at \url{https://github.com/xfey/NTIL}.

\end{abstract}

\section{Introduction}\label{sec:intro}

In recent years, sequence generation has become a crucial approach for implementing a broad range of AI applications, including visual question answering~\citep{wang-etal-2024-soft-knowledge, reich-schultz-2024-uncovering, fan-etal-2024-muffin,liu-etal-2024-aligning,he2025enhancing,wang2025vision,lu2024bounding}, key information extraction~\citep{kim-etal-2024-verifiner,yu-etal-2024-kieval,kang-etal-2024-guidance,wang-etal-2024-order,yi2025score,lu2023punifiedner,lu2024padellm}, object detection~\citep{wen-etal-2024-transitive}, math reasoning~\cite{zhao-etal-2024-docmath}, text spotting~\citep{li-etal-2024-mage,yu2025eve}, and automatic audio recognition~\citep{zhou-etal-2024-copyne}. 




\begin{figure}[t]
  \centering
  \includegraphics[width=0.85\linewidth]{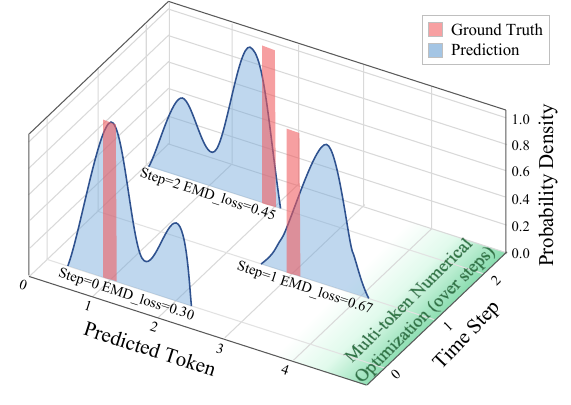}
  \caption{Sequence-level digit token loss illustration.}
  \label{fig:main_fig}
\end{figure}

\begin{figure}[ht]
  \centering
  \includegraphics[width=0.8\linewidth]{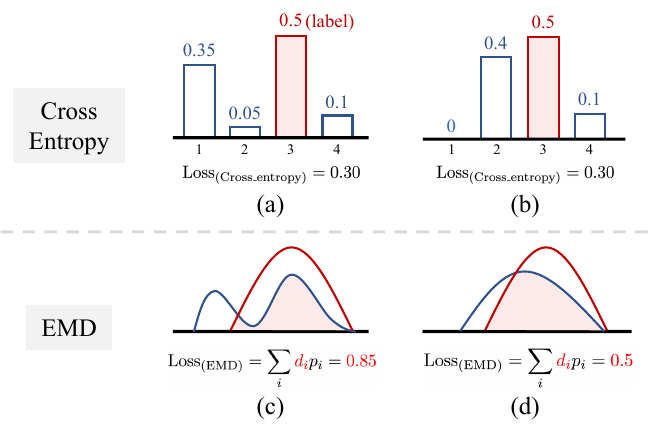}
  \caption{Cross-entropy fails to distinguish predictions, whereas EMD correlates smaller loss for better predicted distributions.}
  \label{fig:emd_example}
  \vspace{-5pt}
\end{figure}

Autoregressive models, especially large language models (LLMs) such as GPT~\cite{achiam2023gpt}, LLaMA~\cite{touvron2023llama,dubey2024llama}, Qwen~\cite{yang2024qwen2,wang2024qwen2} series, with multi-modal large language models (MLLMs) based on them, now dominate the sequence generation tasks. 
During training, these models generate sequences token-by-token, typically using cross-entropy (CE) loss, to minimize the negative log-likelihood of the ground truth token at each time step. 
However, CE loss has several inherent limitations when predicting numerical values.
Specifically, CE suffers from \textbf{Limitation 1} that \textit{it ignores the inherent closeness between numerical tokens, where each digit in a numerical prediction is not independent but related to its neighboring digits. } 
For example, in Figure~\ref{fig:emd_example}(a) and~\ref{fig:emd_example}(b), for the ground truth token ``3'', the CE loss yields same values of $-log(0.5)$ for different prediction distributions. 
However, the distribution in Figure~\ref{fig:emd_example}(b) is more accurate, as it assigns higher probability to the neighboring token ``2''. 



CE also suffers from \textbf{Limitation 2} that \textit{it fails to capture the holistic numerical error when sequential tokens are involved, as it focuses on the precision of each token rather than the overall value}. 
In an autoregressive generation manner, producing a numerical value typically requires consecutive time steps. 
For example, the target value \textit{``0.98''} requires the prediction of four sequential tokens ---  \textit{``0''},  \textit{``.''},  \textit{``9''}, \textit{``8''}. Thus, a prediction such as \textit{1.01} (\textit{``1''},\textit{``.''},\textit{``0''},\textit{``1''}) incurs a high CE loss as the first, third and fourth tokens are significantly different from the target tokens. Conversely, a prediction like \textit{1.98} (\textit{``1''},\textit{``.''},\textit{``9''},\textit{``8''}) could yield a lower CE loss due to a closer match at the token level, despite the overall numerical difference being larger (1.00 vs. 0.03). This discrepancy shows the limitation of CE in evaluating predictions holistically.

To overcome the above issues, we introduce a novel sequence-level numerical prediction loss: \textit{\textbf{N}umerical \textbf{T}oken \textbf{I}ntegrity \textbf{L}oss (NTIL)}. At the token-level, NTIL replaces the traditional CE loss with Earth Mover’s Distance (EMD)~\citep{710701}. Additionally, we enhance the EMD with an \textit{Exponential Position-based Weighting} scheme (Section~\ref{subsec:ebw}), which leverages place-value number systems to better capture the nuanced differences between numerical distributions at each time step.
At the sequence-level, NTIL evaluates the overall numerical difference between predicted and actual sequences through \textit{Multi-token Numerical Optimization} (Section~\ref{sec:method_3.2}), considering all time steps holistically, as illustrated in Figure~\ref{fig:main_fig}. 
It enables NTIL to effectively model the actual value of digit sequences, and capture discrepancies across the consecutive numerical range, moving beyond simple token-by-token comparison.

To the best of the authors' knowledge, it is the first time that EMD is used as an optimization method for autoregressive models. Moreover, our holistic approach is the first of its kind to improve sequential numerical prediction by considering numerical tokens across multiple time steps. 
Our method can be seamlessly integrated into both LLMs and MLLMs. Experimental results show that NTIL boosts performance in tasks requiring precise numerical outputs, such as object detection, text spotting, and math reasoning (Section~\ref{sec:exp}).

\section{Related Work}
\label{sec:prelim_2.3}

The Earth Mover's Distance (EMD) measures the minimal cost of transforming one distribution into another, and has become a valuable metric in deep learning applications. Notably, Wasserstein GAN~\cite{arjovsky2017wasserstein} uses EMD as its loss function to stabilize training in GANs. \citet{cuturi2013sinkhorn} and ~\citet{courty2016optimal} also adopted EMD for smoothing the training procedure. Despite the success of EMD, it has not been applied to autoregressive models. Most recently, autoregressive models, especially LLMs, have advanced NLP~\citep{radford2019language,touvron2023llama,lu-etal-2023-makes,cui2025multi}, and multi-modal tasks~\citep{alayrac2022flamingo,wang2024qwen2,feng2025dolphin,lu2025prolonged}. While the tasks mentioned above require high precision in numerical value prediction, none of the previous works have specifically optimized for this criterion. Our work addresses this gap by focusing on advancing the sequential numerical prediction for autoregressive models.




\section{Method}
\label{sec:method}

This section details the components of the proposed method. Section~\ref{subsec:ebw} proposes exponential weighted EMD to single digits; Section~\ref{sec:method_3.2} describes how we go through multiple digital tokens to derive a simple yet effective numerical measure.


\subsection{Exponential Position-Based Weighting} \label{subsec:ebw} For token-level prediction, to address \textbf{Limitation 1} in Section~\ref{sec:intro}, we replace the conventional CE loss with EMD to account for the ordinal relationship during optimization. The preliminaries for both CE and EMD objectives, and the simplification via numerical prediction, are outlined in Appendix~\ref{sec:preliminaries}.

Furthermore, we extend EMD to account for the place-value number systems, where leading digits have greater numerical significance. 
We implement an exponential weighting scheme to progressively assign weights based on digit positions, to scale their contributions to the loss accordingly:
\begin{equation}
\mathbf{W_{exp}}=\left[(1+\sigma)^{n-i-1}\right]_{i=0}^{n-1},
\end{equation}
where $\sigma$ is the exponential increment rate, and $n$ is the length of consecutive digits. This implementation helps the model understand the order relationship between consecutive numbers.

\subsection{Multi-Token Numerical Optimization}\label{sec:method_3.2}
\begin{figure}[ht]
  \centering
  \includegraphics[width=0.8\linewidth]{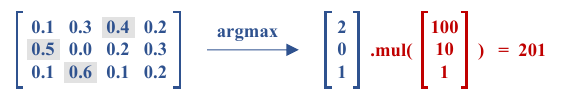}
  \caption{Constructing a numerical value from tokens.}
  \label{fig:gumbel_softmax}
\end{figure}

To overcome \textbf{Limitation 2} outlined in Section~\ref{sec:intro}, we propose the following procedure and losses.
\noindent\textbf{Differentiable Numerical Value Construction.} 
In this step, we construct the complete numerical value from consecutive discrete digital tokens.
Figure~\ref{fig:gumbel_softmax} illustrates how we obtain the digit index from the predicted distribution using $\operatorname{argmax}$ to derive the integer representation. To maintain differentiability, we employ the $\operatorname{Gumbel-softmax}$ approximation with reduced temperature and noise parameters to ensure consistent results.
The resulting tensor is element-wise multiplied with positional indices, scaled by the appropriate powers of 10, and aggregated to obtain the final value. For further implementation details, see Appendix~\ref{appendix:gumbel_softmax}.

\noindent\textbf{Relative Deviation Metric.} For numerical comparison, while absolute difference provides a straightforward measure equivalent to L1 loss, we propose a normalized metric defined as:
\begin{equation}
\mathcal{L}_{relative} = \frac{\left|X-Y\right|}{\operatorname{max}\left(X, Y\right) + \epsilon},
\end{equation}
where $X$ is the sequence-level numerical prediction (\textit{e.g.}, ``234'') and $Y$ is the ground truth, and $\epsilon$ is a small quantity to avoid division by zero. This normalization ensures consistency across different magnitude ranges.


\noindent\textbf{Magnitude Deviation Metric}
We also apply a normalized metric on the order of magnitude as:
\begin{equation}
\mathcal{L}_{magnitude} = \operatorname{log} \left( \frac{\operatorname{max}(X,Y)}{\operatorname{min}(X,Y)} \right).
\end{equation}

The objective penalizes the difference in the order of magnitude between two values. For example, given the pairs $(1, 10)$ and $(1, 100)$, which have similar $\mathcal{L}_{relatvie}$ values $0.90$ and $0.99$, but differ in $\mathcal{L}_{magnitude}$ value: $\log \left( \frac{10}{1} \right) \approx 2.30$ for the first pair and $\log \left( \frac{100}{1} \right) \approx 4.61$ for the second. This results in a larger penalty for greater differences in magnitude. 
The final formulation of NTIL combines the above loss functions, with tunable hyperparameters to weight their individual contributions.
\begin{equation}
\mathcal{L} = \mathbf{W_{exp}}\operatorname{EMD} + \alpha \cdot \mathcal{L}_{relative} + \beta \cdot \mathcal{L}_{magnitude}
\end{equation}





\begin{table*}[t]
\small
\centering
\setlength{\tabcolsep}{7pt}
\scriptsize
\begin{tabular}{llcccccccccc}
\hline
&  & \multicolumn{3}{c}{RefCOCO} & \multicolumn{3}{c}{RefCOCO+} & \multicolumn{2}{c}{RefCOCOg} & \\ \cmidrule(r){3-5} \cmidrule(r){6-8} \cmidrule(r){9-10}
\multirow{-3}{*}{Model} & \multirow{-3}{*}{Method} & Val & TestA & TestB & Val & TestA & TestB & Val & Test  & \multirow{-2}{*}{Avg} \\ \hline
 & CE & 0.839 & 0.865 & \multicolumn{1}{c|}{0.784} & 0.740 &0.797& \multicolumn{1}{c|}{0.664} & 0.792 & \multicolumn{1}{c|}{0.797} & 0.785\\
 & EMD & 0.841  & 0.864 & \multicolumn{1}{c|}{\textbf{0.796}} & 0.749 & 0.805 & \multicolumn{1}{c|}{0.669} & 0.789 & \multicolumn{1}{c|}{0.799} & 0.789 \\
\multirow{-3}{*}{\begin{tabular}[c]{@{}l@{}}PaliGemma\\ (3b)~\cite{beyer2024paligemma}\end{tabular}} & \textbf{Ours} & \textbf{0.844} &  \textbf{0.873} & \multicolumn{1}{c|}{0.791} & \textbf{0.750} & \textbf{0.812} &\multicolumn{1}{c|}{\textbf{0.678}} & \textbf{0.804} & \multicolumn{1}{c|}{\textbf{0.802}} & \textbf{0.795} \\ \hline

 & CE & 0.855 & 0.880 & \multicolumn{1}{c|}{0.813} & \textbf{0.801} &0.843& \multicolumn{1}{c|}{0.741} & 0.799 & \multicolumn{1}{c|}{0.816} &0.818\\
 & EMD & 0.856 & 0.879 & \multicolumn{1}{c|}{\textbf{0.822}} & 0.798 &0.845& \multicolumn{1}{c|}{0.743} & 0.798 & \multicolumn{1}{c|}{0.816} & 0.820\\
\multirow{-3}{*}{\begin{tabular}[c]{@{}l@{}}LLaVA-1.5\\ (7b)~\cite{liu2024improved}\end{tabular}} & \textbf{Ours} & \textbf{0.858} &  \textbf{0.885} & \multicolumn{1}{c|}{0.815} & 0.800 & \textbf{0.853} &\multicolumn{1}{c|}{\textbf{0.747}} & \textbf{0.802} & \multicolumn{1}{c|}{\textbf{0.817}} & \textbf{0.822} \\ \hline
& CE & 0.767 & 0.796 & \multicolumn{1}{c|}{0.734} & 0.706 &0.757& \multicolumn{1}{c|}{0.651} & 0.722 & \multicolumn{1}{c|}{0.731} &0.733\\
& EMD & \textbf{0.779} & 0.805 & \multicolumn{1}{c|}{0.738} & \textbf{0.719} &0.762& \multicolumn{1}{c|}{0.657} & 0.721& \multicolumn{1}{c|}{0.737} &0.740\\
\multirow{-3}{*}{\begin{tabular}[c]{@{}l@{}}Yi-VL\\ (6b)~\cite{young2024yi}\end{tabular}}& \textbf{Ours} & 0.777 &  \textbf{0.808} & \multicolumn{1}{c|}{\textbf{0.741}} & 0.717 & \textbf{0.770} &\multicolumn{1}{c|}{\textbf{0.665}} & \textbf{0.727} & \multicolumn{1}{c|}{\textbf{0.743}} & \textbf{0.744} \\ \hline
 & CE & 0.897 & 0.928 & \multicolumn{1}{c|}{\textbf{0.850}} & 0.841 &\textbf{0.896}& \multicolumn{1}{c|}{0.776} & 0.851 & \multicolumn{1}{c|}{\textbf{0.867}} & 0.863\\
 & EMD & 0.889 & 0.931 & \multicolumn{1}{c|}{0.843} & 0.838 &0.889& \multicolumn{1}{c|}{0.772} & 0.853 & \multicolumn{1}{c|}{0.858} & 0.859\\
\multirow{-3}{*}{\begin{tabular}[c]{@{}l@{}}Qwen2-VL\\ (2b)~\cite{wang2024qwen2}\end{tabular}} & \textbf{Ours} & \textbf{0.898} &  \textbf{0.932} & \multicolumn{1}{c|}{0.849} & \textbf{0.844} & 0.891 &\multicolumn{1}{c|}{\textbf{0.788}} & \textbf{0.858} & \multicolumn{1}{c|}{0.863} & \textbf{0.866} \\ \hline
 & CE & \textbf{0.892} & 0.929 & \multicolumn{1}{c|}{0.841} & 0.842 &0.902& \multicolumn{1}{c|}{0.784} & 0.843 & \multicolumn{1}{c|}{0.848} &0.860\\
& EMD & 0.886 & 0.926 & \multicolumn{1}{c|}{0.834} & 0.843 &0.901& \multicolumn{1}{c|}{0.768} &0.836& \multicolumn{1}{c|}{0.843} &0.855\\
\multirow{-3}{*}{\begin{tabular}[c]{@{}l@{}}Qwen2-VL\\ (7b)~\cite{wang2024qwen2}\end{tabular}} & \textbf{Ours} & 0.889 &  \textbf{0.931} & \multicolumn{1}{c|}{\textbf{0.840}} & \textbf{0.844} & \textbf{0.904} &\multicolumn{1}{c|}{\textbf{0.786}} & \textbf{0.848} & \multicolumn{1}{c|}{\textbf{0.853}} & \textbf{0.862} \\ \hline
\end{tabular}
\caption{Performance comparison (Acc@0.5) of models on image grounding tasks.}
\label{tab:image_grounding}
\end{table*}

\begin{table*}[ht]
\setlength{\tabcolsep}{5pt}
\scriptsize
\begin{minipage}[t]{.56\textwidth}
    \raggedright
    \begin{tabular}{llccccc}
    \hline
     &  & \multicolumn{5}{c}{Dataset} \\ \cmidrule(r){3-6}
    \multirow{-2}{*}{Model} & \multirow{-2}{*}{Method} & CTW1500 & ICDAR1500 & TD500 & Total-Text  & \multirow{-2}{*}{Avg} \\ \hline
     & CE & 0.220 & 0.129 & 0.183 & \multicolumn{1}{c|}{0.259} & 0.193 \\
      & EMD & 0.314 & 0.124 & 0.252 & \multicolumn{1}{c|}{0.307} & 0.241  \\
    \multirow{-3}{*}{\begin{tabular}[c]{@{}l@{}}PaliGemma\\ (3b)\end{tabular}} & \textbf{Ours} & \textbf{0.369} & \textbf{0.155} & \textbf{0.257} & \multicolumn{1}{c|}{\textbf{0.318}} & \textbf{0.263} \\ \hline
    
     & CE & \textbf{0.682} & 0.370 & 0.753 & \multicolumn{1}{c|}{0.673} & 0.586 \\
       & EMD & 0.668 & 0.398 & \textbf{0.778} & \multicolumn{1}{c|}{0.678} & 0.594   \\
    \multirow{-3}{*}{\begin{tabular}[c]{@{}l@{}}Yi-VL\\ (6b)\end{tabular}} & \textbf{Ours} & 0.680 & \textbf{0.403} & 0.752 & \multicolumn{1}{c|}{\textbf{0.678}} & \textbf{0.597} \\ \hline
    
     & CE & 0.786 & 0.538 & 0.851 & \multicolumn{1}{c|}{0.827} & 0.720 \\
       & EMD & \textbf{0.786} & 0.535 & \textbf{0.867} & \multicolumn{1}{c|}{0.808} & 0.718  \\
    \multirow{-3}{*}{\begin{tabular}[c]{@{}l@{}}Qwen2-VL\\ (2b)\end{tabular}} & \textbf{Ours} & 0.776 & \textbf{0.577} & 0.854 & \multicolumn{1}{c|}{\textbf{0.835}} & \textbf{0.732} \\ \hline
    
     & CE & \textbf{0.771} & 0.648 & \textbf{0.889} & \multicolumn{1}{c|}{0.864} & 0.764 \\
       & EMD & 0.762 & 0.625 & 0.874 & \multicolumn{1}{c|}{0.860} & 0.751  \\
    \multirow{-3}{*}{\begin{tabular}[c]{@{}l@{}}Qwen2-VL\\ (7b)\end{tabular}} & \textbf{Ours} & 0.770 & \textbf{0.669} & 0.869 & \multicolumn{1}{c|}{\textbf{0.872}} & \textbf{0.770} \\ \hline
     & CE & 0.735 & 0.490 & 0.821 & \multicolumn{1}{c|}{0.786} & 0.675 \\
       & EMD & 0.724 & 0.545 & \textbf{0.840} & \multicolumn{1}{c|}{0.776} & 0.690  \\
    \multirow{-3}{*}{\begin{tabular}[c]{@{}l@{}}LLaVA-1.5\\ (7b)\end{tabular}} & \textbf{Ours} & \textbf{0.739} & \textbf{0.547} & 0.839 & \multicolumn{1}{c|}{\textbf{0.791}} & \textbf{0.698} \\ \hline
    
    \end{tabular}
    \captionsetup{justification=raggedright,singlelinecheck=false}
    \caption{Performance (Acc@0.5) on scene text detection tasks.}
    \label{tab:text_detection}
\end{minipage}%
\renewcommand{\arraystretch}{1.23}
\begin{minipage}[t]{.43\textwidth}
    \centering
    \begin{tabular}{lccc}
    \hline
    \multirow{2}{*}{Model} & \multicolumn{3}{c}{Accuracy (\%)} \\ \cline{2-4}
    & CE & EMD & \textbf{Ours} \\ \hline
    \begin{tabular}[c]{@{}l@{}}Baichuan2 (7b)\\~\cite{yang2023baichuan}\end{tabular} & 44.3 & 46.6 & \textbf{46.9} \\ \hline
    \begin{tabular}[c]{@{}l@{}}Qwen2.5 (1.5b)\\~\cite{qwen2.5}\end{tabular} & 40.3 & 40.7 & \textbf{42.4} \\ \hline
    \begin{tabular}[c]{@{}l@{}}LLaMA3 (8b)\\~\cite{dubey2024llama}\end{tabular} & 61.9 & 61.8 & \textbf{61.9} \\ \hline
    \begin{tabular}[c]{@{}l@{}}Yi (6b)\\~\cite{young2024yi}\end{tabular} & 53.0 & \textbf{54.6} & 54.4 \\ \hline
    \begin{tabular}[c]{@{}l@{}}MiniCPM3 (4b)\\~\cite{hu2024minicpm}\end{tabular} & 66.8 & 68.2 & \textbf{68.6} \\ \hline
    \end{tabular}
    \captionsetup{width=.86\textwidth}
    \caption{Performance comparison of accuracies on the arithmetic calculation task.}
    \label{tab:arithmetic_calculation}
\end{minipage}
\vspace{-3ex}
\end{table*}

\begin{table}[t]
\small
\centering
\scriptsize
\resizebox{\columnwidth}{!}{%
\begin{tabular}{llcccc}
\hline
\multirow{-1}{*}{Metric} & \multirow{-1}{*}{Method} & \begin{tabular}[c]{@{}c@{}}LLaVA-1.5 \\ (7b)\end{tabular} & \begin{tabular}[c]{@{}c@{}}Qwen2-VL\\ (7b)\end{tabular} & \begin{tabular}[c]{@{}c@{}}Qwen2-VL\\ (2b)\end{tabular} & \begin{tabular}[c]{@{}c@{}}Yi-VL\\ (6b) \end{tabular}  \\ \hline
 & CE & 95.1 & 75.0 & 81.3 & 76.2 \\
 & EMD & 95.3 & 78.7 & 81.7 & 75.1 \\
\multirow{-3}{*}{\begin{tabular}[c]{@{}l@{}}Accuracy\\ (\%) $\uparrow$ \end{tabular}} & \textbf{Ours} & \textbf{98.3} & \textbf{80.5} & \textbf{85.3} & \textbf{87.4} \\ \hline
 & CE & 8.52 & 30.84 & 32.34 & 56.58 \\
 & EMD & 7.98 & 30.78 & 31.98 & 54.78 \\
\multirow{-3}{*}{\begin{tabular}[c]{@{}l@{}}Time gap\\ (minute) $\downarrow$ \end{tabular}} & \textbf{Ours} & \textbf{4.14} & \textbf{27.72} & \textbf{24.66} & \textbf{26.58} \\ \hline
\end{tabular}
}
\caption{Performance of the clock time recognition task.}
\label{tab:time_clock}
\end{table}

\begin{table}[t]
\small
\centering
\scriptsize
\resizebox{\columnwidth}{!}{%
\begin{tabular}{llccccc}
\hline
Dataset & Method & \begin{tabular}[c]{@{}c@{}}Qwen2-vl\\ (2b)\end{tabular} & \begin{tabular}[c]{@{}c@{}}Qwen2-vl\\ (7b)\end{tabular} & \begin{tabular}[c]{@{}c@{}}LLaVA-1.5\\ (7b)\end{tabular} & \begin{tabular}[c]{@{}c@{}}Yi-VL\\ (6b)\end{tabular} & \begin{tabular}[c]{@{}c@{}}PaliGemma\\ (3b)\end{tabular} \\ \hline
 & CE & 0.139 & 0.184 & 0.146 & 0.143 & 0.097 \\
 & EMD & 0.130 & 0.188 & \textbf{0.148} & 0.142 & 0.088 \\
\multirow{-3}{*}{Mathvision} & \textbf{Ours} & \textbf{0.145} & \textbf{0.191} & 0.146 & \textbf{0.153} & \textbf{0.098} \\ \hline
 & CE & 0.248 & \textbf{0.315} & 0.140 & 0.187 & 0.143 \\
 & EMD & \textbf{0.262} & 0.303 & 0.157 & 0.192 & 0.149 \\
\multirow{-3}{*}{Mathvista} & \textbf{Ours} & 0.251 & 0.300 & \textbf{0.170} & \textbf{0.222} & \textbf{0.157} \\ \hline
\end{tabular}
}
\caption{Performance of the math reasoning task.}
\label{tab:math_reasoning}
\end{table}


\begin{table}[t]
\centering
\setlength{\tabcolsep}{4pt}
\scalebox{0.5}{\begin{tabular}{ccc|cc|cc|cc|cc}
\toprule
~ &~&~& \multicolumn{2}{c|}{\textbf{PaliGemma}} & \multicolumn{2}{c|}{\textbf{LLaVA-1.5}} & \multicolumn{2}{c|}{\textbf{Qwen2-VL}} & \multicolumn{2}{c}{\textbf{Yi-VL}} \\\midrule
 Exp & Rel & Mag & Mathvision & Mathvista & Mathvision & Mathvista & \multicolumn{2}{c|}{Clock\_Time} & \multicolumn{2}{c}{Clock\_Time}  \\    \midrule
$\times$ & \checkmark & \checkmark &0.096&0.137&0.151&0.166& \multicolumn{2}{c|}{0.798}& \multicolumn{2}{c}{0.834} \\
\checkmark & $\times$ & \checkmark &0.095&0.137&0.145&0.154& \multicolumn{2}{c|}{0.790}& \multicolumn{2}{c}{0.856} \\
\checkmark & \checkmark & $\times$ &0.094&0.142&\textbf{0.160}&0.143& \multicolumn{2}{c|}{0.816}& \multicolumn{2}{c}{\textbf{0.876}} \\
\checkmark & \checkmark & \checkmark &\textbf{0.098}&\textbf{0.157}& 0.146 & \textbf{0.170}& \multicolumn{2}{c|}{\textbf{0.853}}& \multicolumn{2}{c}{0.874} \\
\bottomrule
\end{tabular}
}
\caption{Ablations on NTIL. Exp: Exponential Position-Based Weighting. REL: Relative Deviation Metric. Mag: Magnitude Deviation Metric.}
\label{tab:ablation}
\end{table}

\begin{figure}[ht]
  \centering
  \includegraphics[width=0.9\linewidth]{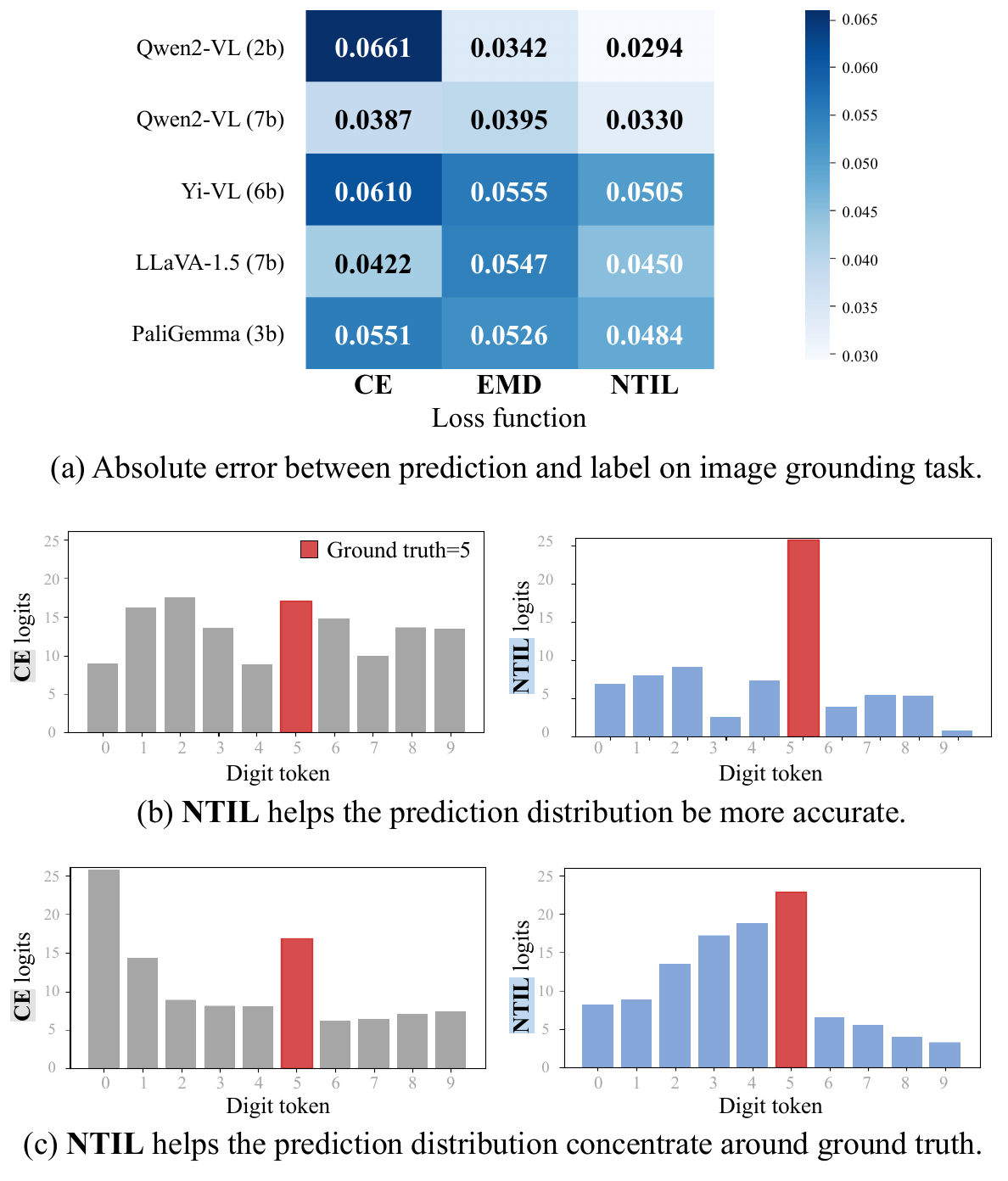}
  \caption{Results for quantitative analysis.}
  \label{fig:quantitative}
\end{figure}
\section{Experiments and Results}
\label{sec:exp}

This section presents a comprehensive empirical evaluation of the proposed NTIL across various LLMs/MLLMs (Section~\ref{sec:exp_4.1}). \textbf{CE}~\citep{shannon1948mathematical} and \textbf{EMD}~\citep{710701} are chosen as baselines due to their widespread adoption. The evaluation encompasses multiple task domains that focus on numerical prediction including \textit{Image Grounding}, \textit{Scene Text Detection}, \textit{Clock Time Recognition}, \textit{Mathematical Reasoning} and \textit{Arithmetic Calculations}. 
Appendix~\ref{appendix:data} provides details on tasks, datasets, and evaluation metrics.
We also conduct systematic ablation studies to evaluate the critical components of our approach. Implementation details are available in Appendix~\ref{appendix:imple}.

\subsection{Main Results}
\label{sec:exp_4.1}


\subsection{Results of MLLMs}

\noindent \textbf{Image Grounding} As shown in Table~\ref{tab:image_grounding}, our method outperforms both CE and EMD across nearly all datasets and VLM backbones, as evidenced by the overall performance improvements. 

\noindent \textbf{Scene Text Detection} Table~\ref{tab:text_detection} shows that our method improves accuracy across multiple datasets, demonstrating its effectiveness in predicting multiple object coordinates.

\noindent \textbf{Clock Time Recognition} Table~\ref{tab:time_clock} demonstrates that NTIL surpasses CE and EMD significantly in performance across all model architectures.


\noindent \textbf{Mathematical Reasoning} As shown in Table~\ref{tab:math_reasoning}, our method outperforms CE and EMD across all datasets, with the most significant improvements seen in the Mathvision dataset using the Qwen2-VL (2b) and in Mathvista with the Yi-VL (6b). 

\subsection{Results of LLMs}
\noindent \textbf{Arithmetic Calculation} 
As shown in Table~\ref{tab:arithmetic_calculation}, our method improves accuracy across multiple LLMs, though LLaMA3 shows minimal gains, possibly due to its extensive pre-training. 
Overall, the majority of cases show that for numerical predictions, while EMD performs comparably or marginally better than CE loss, NTIL consistently delivers superior results in most scenarios. This underscores the effectiveness and generalizability of NTIL.


\subsection{Ablation Analysis}

Table~\ref{tab:ablation} indicates that incorporating all components of NTIL generally leads to better performance, as evidenced by the highest scores in most metrics when all components are enabled. 
As an exception, the inclusion of Magnitude leads to worse results in Mathvision for LLaVA-1.5, which indicates the fluctuation of applying Magnitude in some cases.

\subsection{Quantitative Analysis}

As shown in Figure~\ref{fig:quantitative}(a), NTIL achieves the lowest absolute errors among all models, indicating more consistent performance compared to CE and EMD. 
Figure~\ref{fig:quantitative}(b) and~\ref{fig:quantitative}(c) illustrate that NTIL produces more accurate predictions with distributions more concentrated around the ground truth.
Overall, NITL offers more stability and lower variability. Qualitative examples can be seen in Appendix~\ref{appendix:quali}.


\section{Conclusion}

We propose NTIL, which improves numerical prediction accuracy in LLMs at both the token and sequence levels. Experiments show improvement across multiple datasets and models, highlighting effectiveness of NTIL.

\section*{Limitations}

The limitations of the NTIL include the exponential position-based weighting scheme, while effective in many cases, 
had limited or negative impact in certain configurations, such as with the Mathvision dataset 
and LLaVA-1.5 model. Future exploration could focus on refining the exponential position-based 
weighting scheme with adaptive strategies to address its inconsistent impact. 

Furthermore, NTIL introduces additional computational overhead compared to CE loss, 
resulting in reduced training efficiency. This trade-off between performance improvement 
and computational cost needs to be considered in practical applications.

Another limitation arises from the tokenization strategies of certain models like LLaMA-3, 
which encode common multi-digit numbers (e.g., "123") as a single token. Such cases require 
special handling in NTIL's implementation, adding complexity to the framework.



\bibliography{custom}

\begin{thebibliography}{56}
\providecommand{\natexlab}[1]{#1}

\bibitem[{Achiam et~al.(2023)Achiam, Adler, Agarwal, Ahmad, Akkaya, Aleman,
  Almeida, Altenschmidt, Altman, Anadkat et~al.}]{achiam2023gpt}
Josh Achiam, Steven Adler, Sandhini Agarwal, Lama Ahmad, Ilge Akkaya,
  Florencia~Leoni Aleman, Diogo Almeida, Janko Altenschmidt, Sam Altman,
  Shyamal Anadkat, et~al. 2023.
\newblock Gpt-4 technical report.
\newblock \emph{arXiv preprint arXiv:2303.08774}.

\bibitem[{Alayrac et~al.(2022)Alayrac, Donahue, Luc, Miech, Barr, Hasson, Lenc,
  Mensch, Millican, Reynolds et~al.}]{alayrac2022flamingo}
Jean-Baptiste Alayrac, Jeff Donahue, Pauline Luc, Antoine Miech, Iain Barr,
  Yana Hasson, Karel Lenc, Arthur Mensch, Katherine Millican, Malcolm Reynolds,
  et~al. 2022.
\newblock Flamingo: a visual language model for few-shot learning.
\newblock \emph{Advances in neural information processing systems},
  35:23716--23736.

\bibitem[{Arjovsky et~al.(2017)Arjovsky, Chintala, and
  Bottou}]{arjovsky2017wasserstein}
Martin Arjovsky, S~Chintala, and L{\'e}on Bottou. 2017.
\newblock Wasserstein gan. arxiv preprint arxiv: 170107875.
\newblock \emph{arXiv preprint arXiv:1701.07875}.

\bibitem[{Beyer et~al.(2024)Beyer, Steiner, Pinto, Kolesnikov, Wang, Salz,
  Neumann, Alabdulmohsin, Tschannen, Bugliarello et~al.}]{beyer2024paligemma}
Lucas Beyer, Andreas Steiner, Andr{\'e}~Susano Pinto, Alexander Kolesnikov,
  Xiao Wang, Daniel Salz, Maxim Neumann, Ibrahim Alabdulmohsin, Michael
  Tschannen, Emanuele Bugliarello, et~al. 2024.
\newblock Paligemma: A versatile 3b vlm for transfer.
\newblock \emph{arXiv preprint arXiv:2407.07726}.

\bibitem[{Ch'ng and Chan(2017)}]{ch2017total}
Chee~Kheng Ch'ng and Chee~Seng Chan. 2017.
\newblock Total-text: A comprehensive dataset for scene text detection and
  recognition.
\newblock In \emph{2017 14th IAPR international conference on document analysis
  and recognition (ICDAR)}, volume~1, pages 935--942. IEEE.

\bibitem[{Courty et~al.(2016)Courty, Flamary, Tuia, and
  Rakotomamonjy}]{courty2016optimal}
Nicolas Courty, R{\'e}mi Flamary, Devis Tuia, and Alain Rakotomamonjy. 2016.
\newblock Optimal transport for domain adaptation.
\newblock \emph{IEEE transactions on pattern analysis and machine
  intelligence}, 39(9):1853--1865.

\bibitem[{Cui et~al.(2025)Cui, Zhu, Qin, Xie, Zhou, and Li}]{cui2025multi}
Xiao Cui, Mo~Zhu, Yulei Qin, Liang Xie, Wengang Zhou, and Houqiang Li. 2025.
\newblock Multi-level optimal transport for universal cross-tokenizer knowledge
  distillation on language models.
\newblock In \emph{Proceedings of the AAAI Conference on Artificial
  Intelligence}, volume~39, pages 23724--23732.

\bibitem[{Cuturi(2013)}]{cuturi2013sinkhorn}
Marco Cuturi. 2013.
\newblock Sinkhorn distances: Lightspeed computation of optimal transport.
\newblock \emph{Advances in neural information processing systems}, 26.

\bibitem[{Dubey et~al.(2024)Dubey, Jauhri, Pandey, Kadian, Al-Dahle, Letman,
  Mathur, Schelten, Yang, Fan et~al.}]{dubey2024llama}
Abhimanyu Dubey, Abhinav Jauhri, Abhinav Pandey, Abhishek Kadian, Ahmad
  Al-Dahle, Aiesha Letman, Akhil Mathur, Alan Schelten, Amy Yang, Angela Fan,
  et~al. 2024.
\newblock The llama 3 herd of models.
\newblock \emph{arXiv preprint arXiv:2407.21783}.

\bibitem[{Fan et~al.(2024)Fan, Gu, Zhou, Yan, Jiang, Kuo, Zhao, Guan, and
  Wang}]{fan-etal-2024-muffin}
Yue Fan, Jing Gu, Kaiwen Zhou, Qianqi Yan, Shan Jiang, Ching-Chen Kuo, Yang
  Zhao, Xinze Guan, and Xin Wang. 2024.
\newblock \href {https://doi.org/10.18653/v1/2024.acl-long.370} {Muffin or
  {C}hihuahua? challenging multimodal large language models with multipanel
  {VQA}}.
\newblock In \emph{Proceedings of the 62nd Annual Meeting of the Association
  for Computational Linguistics (Volume 1: Long Papers)}, pages 6845--6863,
  Bangkok, Thailand. Association for Computational Linguistics.

\bibitem[{Feng et~al.(2025)Feng, Wei, Fei, Shi, Han, Liao, Lu, Wu, Liu, Lin
  et~al.}]{feng2025dolphin}
Hao Feng, Shu Wei, Xiang Fei, Wei Shi, Yingdong Han, Lei Liao, Jinghui Lu,
  Binghong Wu, Qi~Liu, Chunhui Lin, et~al. 2025.
\newblock Dolphin: Document image parsing via heterogeneous anchor prompting.
\newblock \emph{arXiv preprint arXiv:2505.14059}.

\bibitem[{gpiosenka(2022)}]{time_dataset_kaggle}
gpiosenka. 2022.
\newblock \href
  {https://www.kaggle.com/datasets/gpiosenka/time-image-datasetclassification}
  {Time-image dataset-classification}.

\bibitem[{He et~al.(2025)He, Wang, Li, Wang, Sun, Yin, Zhang, and
  Wang}]{he2025enhancing}
Yangfan He, Jianhui Wang, Kun Li, Yijin Wang, Li~Sun, Jun Yin, Miao Zhang, and
  Xueqian Wang. 2025.
\newblock Enhancing intent understanding for ambiguous prompts through
  human-machine co-adaptation.
\newblock \emph{arXiv preprint arXiv:2501.15167}.

\bibitem[{Hou et~al.(2016)Hou, Yu, and Samaras}]{hou2016squared}
Le~Hou, Chen-Ping Yu, and Dimitris Samaras. 2016.
\newblock Squared earth mover's distance-based loss for training deep neural
  networks.
\newblock \emph{arXiv preprint arXiv:1611.05916}.

\bibitem[{Hu et~al.(2024)Hu, Tu, Han, He, Cui, Long, Zheng, Fang, Huang, Zhao
  et~al.}]{hu2024minicpm}
Shengding Hu, Yuge Tu, Xu~Han, Chaoqun He, Ganqu Cui, Xiang Long, Zhi Zheng,
  Yewei Fang, Yuxiang Huang, Weilin Zhao, et~al. 2024.
\newblock Minicpm: Unveiling the potential of small language models with
  scalable training strategies.
\newblock \emph{arXiv preprint arXiv:2404.06395}.

\bibitem[{Huijben et~al.(2022)Huijben, Kool, Paulus, and
  Van~Sloun}]{huijben2022review}
Iris~AM Huijben, Wouter Kool, Max~B Paulus, and Ruud~JG Van~Sloun. 2022.
\newblock A review of the gumbel-max trick and its extensions for discrete
  stochasticity in machine learning.
\newblock \emph{IEEE transactions on pattern analysis and machine
  intelligence}, 45(2):1353--1371.

\bibitem[{Jang et~al.(2016)Jang, Gu, and Poole}]{jang2016categorical}
Eric Jang, Shixiang Gu, and Ben Poole. 2016.
\newblock Categorical reparameterization with gumbel-softmax.
\newblock \emph{arXiv preprint arXiv:1611.01144}.

\bibitem[{Kang et~al.(2024)Kang, Seo, Jung, Jung, Chang, and
  Chung}]{kang-etal-2024-guidance}
Hyeonseok Kang, Hyein Seo, Jeesu Jung, Sangkeun Jung, Du-Seong Chang, and Riwoo
  Chung. 2024.
\newblock \href {https://doi.org/10.18653/v1/2024.acl-short.61} {Guidance-based
  prompt data augmentation in specialized domains for named entity
  recognition}.
\newblock In \emph{Proceedings of the 62nd Annual Meeting of the Association
  for Computational Linguistics (Volume 2: Short Papers)}, pages 665--672,
  Bangkok, Thailand. Association for Computational Linguistics.

\bibitem[{Karatzas et~al.(2015)Karatzas, Gomez-Bigorda, Nicolaou, Ghosh,
  Bagdanov, Iwamura, Matas, Neumann, Chandrasekhar, Lu
  et~al.}]{karatzas2015icdar}
Dimosthenis Karatzas, Lluis Gomez-Bigorda, Anguelos Nicolaou, Suman Ghosh,
  Andrew Bagdanov, Masakazu Iwamura, Jiri Matas, Lukas Neumann,
  Vijay~Ramaseshan Chandrasekhar, Shijian Lu, et~al. 2015.
\newblock Icdar 2015 competition on robust reading.
\newblock In \emph{2015 13th international conference on document analysis and
  recognition (ICDAR)}, pages 1156--1160. IEEE.

\bibitem[{Kim et~al.(2024)Kim, Seo, Chae, Yeo, and
  Lee}]{kim-etal-2024-verifiner}
Seoyeon Kim, Kwangwook Seo, Hyungjoo Chae, Jinyoung Yeo, and Dongha Lee. 2024.
\newblock \href {https://doi.org/10.18653/v1/2024.acl-long.134} {{V}erifi{NER}:
  Verification-augmented {NER} via knowledge-grounded reasoning with large
  language models}.
\newblock In \emph{Proceedings of the 62nd Annual Meeting of the Association
  for Computational Linguistics (Volume 1: Long Papers)}, pages 2441--2461,
  Bangkok, Thailand. Association for Computational Linguistics.

\bibitem[{Li et~al.(2024)Li, Li, Cui, Bi, Wang, Wang, Yang, Shi, and
  Zhang}]{li-etal-2024-mage}
Yafu Li, Qintong Li, Leyang Cui, Wei Bi, Zhilin Wang, Longyue Wang, Linyi Yang,
  Shuming Shi, and Yue Zhang. 2024.
\newblock \href {https://doi.org/10.18653/v1/2024.acl-long.3} {{MAGE}:
  Machine-generated text detection in the wild}.
\newblock In \emph{Proceedings of the 62nd Annual Meeting of the Association
  for Computational Linguistics (Volume 1: Long Papers)}, pages 36--53,
  Bangkok, Thailand. Association for Computational Linguistics.

\bibitem[{Lin et~al.(2014)Lin, Maire, Belongie, Hays, Perona, Ramanan,
  Doll{\'a}r, and Zitnick}]{lin2014microsoft}
Tsung-Yi Lin, Michael Maire, Serge Belongie, James Hays, Pietro Perona, Deva
  Ramanan, Piotr Doll{\'a}r, and C~Lawrence Zitnick. 2014.
\newblock Microsoft coco: Common objects in context.
\newblock In \emph{Computer Vision--ECCV 2014: 13th European Conference,
  Zurich, Switzerland, September 6-12, 2014, Proceedings, Part V 13}, pages
  740--755. Springer.

\bibitem[{Liu et~al.(2024{\natexlab{a}})Liu, Li, Li, and Lee}]{liu2024improved}
Haotian Liu, Chunyuan Li, Yuheng Li, and Yong~Jae Lee. 2024{\natexlab{a}}.
\newblock Improved baselines with visual instruction tuning.
\newblock In \emph{Proceedings of the IEEE/CVF Conference on Computer Vision
  and Pattern Recognition}, pages 26296--26306.

\bibitem[{Liu et~al.(2024{\natexlab{b}})Liu, Wang, Wu, Li, Lv, Ling, JianHao,
  Zhang, Zheng, and Huang}]{liu-etal-2024-aligning}
Wenhao Liu, Xiaohua Wang, Muling Wu, Tianlong Li, Changze Lv, Zixuan Ling, Zhu
  JianHao, Cenyuan Zhang, Xiaoqing Zheng, and Xuanjing Huang.
  2024{\natexlab{b}}.
\newblock \href {https://doi.org/10.18653/v1/2024.acl-long.572} {Aligning large
  language models with human preferences through representation engineering}.
\newblock In \emph{Proceedings of the 62nd Annual Meeting of the Association
  for Computational Linguistics (Volume 1: Long Papers)}, pages 10619--10638,
  Bangkok, Thailand. Association for Computational Linguistics.

\bibitem[{Lu et~al.(2024{\natexlab{a}})Lu, Wang, Yang, Liu, Mac~Namee, and
  Huang}]{lu2024padellm}
Jinghui Lu, Yanjie Wang, Ziwei Yang, Xuejing Liu, Brian Mac~Namee, and Can
  Huang. 2024{\natexlab{a}}.
\newblock Padellm-ner: parallel decoding in large language models for named
  entity recognition.
\newblock \emph{Advances in Neural Information Processing Systems},
  37:117853--117880.

\bibitem[{Lu et~al.(2024{\natexlab{b}})Lu, Yu, Wang, Ye, Tang, Yang, Wu, Liu,
  Feng, Wang et~al.}]{lu2024bounding}
Jinghui Lu, Haiyang Yu, Yanjie Wang, Yongjie Ye, Jingqun Tang, Ziwei Yang,
  Binghong Wu, Qi~Liu, Hao Feng, Han Wang, et~al. 2024{\natexlab{b}}.
\newblock A bounding box is worth one token: Interleaving layout and text in a
  large language model for document understanding.
\newblock \emph{arXiv preprint arXiv:2407.01976}.

\bibitem[{Lu et~al.(2025)Lu, Yu, Xu, Ran, Tang, Wang, Shan, Fu, Feng, Tang
  et~al.}]{lu2025prolonged}
Jinghui Lu, Haiyang Yu, Siliang Xu, Shiwei Ran, Guozhi Tang, Siqi Wang, Bin
  Shan, Teng Fu, Hao Feng, Jingqun Tang, et~al. 2025.
\newblock Prolonged reasoning is not all you need: Certainty-based adaptive
  routing for efficient llm/mllm reasoning.
\newblock \emph{arXiv preprint arXiv:2505.15154}.

\bibitem[{Lu et~al.(2023{\natexlab{a}})Lu, Zhao, Mac~Namee, and
  Tan}]{lu2023punifiedner}
Jinghui Lu, Rui Zhao, Brian Mac~Namee, and Fei Tan. 2023{\natexlab{a}}.
\newblock Punifiedner: A prompting-based unified ner system for diverse
  datasets.
\newblock In \emph{Proceedings of the AAAI conference on artificial
  intelligence}, volume~37, pages 13327--13335.

\bibitem[{Lu et~al.(2023{\natexlab{b}})Lu, Zhu, Han, Zhao, Mac~Namee, and
  Tan}]{lu-etal-2023-makes}
Jinghui Lu, Dongsheng Zhu, Weidong Han, Rui Zhao, Brian Mac~Namee, and Fei Tan.
  2023{\natexlab{b}}.
\newblock \href {https://doi.org/10.18653/v1/2023.acl-long.128} {What makes
  pre-trained language models better zero-shot learners?}
\newblock In \emph{Proceedings of the 61st Annual Meeting of the Association
  for Computational Linguistics (Volume 1: Long Papers)}, pages 2288--2303,
  Toronto, Canada. Association for Computational Linguistics.

\bibitem[{Lu et~al.(2023{\natexlab{c}})Lu, Bansal, Xia, Liu, Li, Hajishirzi,
  Cheng, Chang, Galley, and Gao}]{lu2023mathvista}
Pan Lu, Hritik Bansal, Tony Xia, Jiacheng Liu, Chunyuan Li, Hannaneh
  Hajishirzi, Hao Cheng, Kai-Wei Chang, Michel Galley, and Jianfeng Gao.
  2023{\natexlab{c}}.
\newblock Mathvista: Evaluating mathematical reasoning of foundation models in
  visual contexts.
\newblock \emph{arXiv preprint arXiv:2310.02255}.

\bibitem[{Mao et~al.(2016)Mao, Huang, Toshev, Camburu, Yuille, and
  Murphy}]{mao2016generation}
Junhua Mao, Jonathan Huang, Alexander Toshev, Oana Camburu, Alan~L Yuille, and
  Kevin Murphy. 2016.
\newblock Generation and comprehension of unambiguous object descriptions.
\newblock In \emph{Proceedings of the IEEE conference on computer vision and
  pattern recognition}, pages 11--20.

\bibitem[{Radford et~al.(2019)Radford, Wu, Child, Luan, Amodei, Sutskever
  et~al.}]{radford2019language}
Alec Radford, Jeffrey Wu, Rewon Child, David Luan, Dario Amodei, Ilya
  Sutskever, et~al. 2019.
\newblock Language models are unsupervised multitask learners.
\newblock \emph{OpenAI blog}, 1(8):9.

\bibitem[{Reich and Schultz(2024)}]{reich-schultz-2024-uncovering}
Daniel Reich and Tanja Schultz. 2024.
\newblock \href {https://doi.org/10.18653/v1/2024.acl-long.241} {Uncovering the
  full potential of visual grounding methods in {VQA}}.
\newblock In \emph{Proceedings of the 62nd Annual Meeting of the Association
  for Computational Linguistics (Volume 1: Long Papers)}, pages 4406--4419,
  Bangkok, Thailand. Association for Computational Linguistics.

\bibitem[{Rubner et~al.(1998)Rubner, Tomasi, and Guibas}]{710701}
Y.~Rubner, C.~Tomasi, and L.J. Guibas. 1998.
\newblock \href {https://doi.org/10.1109/ICCV.1998.710701} {A metric for
  distributions with applications to image databases}.
\newblock In \emph{Sixth International Conference on Computer Vision (IEEE Cat.
  No.98CH36271)}, pages 59--66.

\bibitem[{Saxton et~al.(2019)Saxton, Grefenstette, Hill, and
  Kohli}]{saxton2019analysing}
David Saxton, Edward Grefenstette, Felix Hill, and Pushmeet Kohli. 2019.
\newblock Analysing mathematical reasoning abilities of neural models.
\newblock \emph{arXiv preprint arXiv:1904.01557}.

\bibitem[{Shannon(1948)}]{shannon1948mathematical}
Claude~Elwood Shannon. 1948.
\newblock A mathematical theory of communication.
\newblock \emph{The Bell system technical journal}, 27(3):379--423.

\bibitem[{Team(2024)}]{qwen2.5}
Qwen Team. 2024.
\newblock \href {https://qwenlm.github.io/blog/qwen2.5/} {Qwen2.5: A party of
  foundation models}.

\bibitem[{Touvron et~al.(2023)Touvron, Lavril, Izacard, Martinet, Lachaux,
  Lacroix, Rozi{\`e}re, Goyal, Hambro, Azhar et~al.}]{touvron2023llama}
Hugo Touvron, Thibaut Lavril, Gautier Izacard, Xavier Martinet, Marie-Anne
  Lachaux, Timoth{\'e}e Lacroix, Baptiste Rozi{\`e}re, Naman Goyal, Eric
  Hambro, Faisal Azhar, et~al. 2023.
\newblock Llama: Open and efficient foundation language models.
\newblock \emph{arXiv preprint arXiv:2302.13971}.

\bibitem[{Wan et~al.(2020)Wan, Dai, Zhang, He, Tian, Xie, Wu, Yu, Xu, Chen
  et~al.}]{wan2020fbnetv2}
Alvin Wan, Xiaoliang Dai, Peizhao Zhang, Zijian He, Yuandong Tian, Saining Xie,
  Bichen Wu, Matthew Yu, Tao Xu, Kan Chen, et~al. 2020.
\newblock Fbnetv2: Differentiable neural architecture search for spatial and
  channel dimensions.
\newblock In \emph{Proceedings of the IEEE/CVF conference on computer vision
  and pattern recognition}, pages 12965--12974.

\bibitem[{Wang et~al.(2025)Wang, Ye, Li, Nie, Lu, Tang, Wang, and
  Huang}]{wang2025vision}
Han Wang, Yongjie Ye, Bingru Li, Yuxiang Nie, Jinghui Lu, Jingqun Tang, Yanjie
  Wang, and Can Huang. 2025.
\newblock Vision as lora.
\newblock \emph{arXiv preprint arXiv:2503.20680}.

\bibitem[{Wang et~al.(2024{\natexlab{a}})Wang, Cheng, Zhang, Soh, and
  Bing}]{wang-etal-2024-order}
Huiming Wang, Liying Cheng, Wenxuan Zhang, De~Wen Soh, and Lidong Bing.
  2024{\natexlab{a}}.
\newblock \href {https://doi.org/10.18653/v1/2024.acl-long.421} {Order-agnostic
  data augmentation for few-shot named entity recognition}.
\newblock In \emph{Proceedings of the 62nd Annual Meeting of the Association
  for Computational Linguistics (Volume 1: Long Papers)}, pages 7792--7807,
  Bangkok, Thailand. Association for Computational Linguistics.

\bibitem[{Wang et~al.(2024{\natexlab{b}})Wang, Pan, Shi, Lu, Zhan, and
  Li}]{wang2024measuring}
Ke~Wang, Junting Pan, Weikang Shi, Zimu Lu, Mingjie Zhan, and Hongsheng Li.
  2024{\natexlab{b}}.
\newblock Measuring multimodal mathematical reasoning with math-vision dataset.
\newblock \emph{arXiv preprint arXiv:2402.14804}.

\bibitem[{Wang et~al.(2024{\natexlab{c}})Wang, Bai, Tan, Wang, Fan, Bai, Chen,
  Liu, Wang, Ge et~al.}]{wang2024qwen2}
Peng Wang, Shuai Bai, Sinan Tan, Shijie Wang, Zhihao Fan, Jinze Bai, Keqin
  Chen, Xuejing Liu, Jialin Wang, Wenbin Ge, et~al. 2024{\natexlab{c}}.
\newblock Qwen2-vl: Enhancing vision-language model's perception of the world
  at any resolution.
\newblock \emph{arXiv preprint arXiv:2409.12191}.

\bibitem[{Wang et~al.(2024{\natexlab{d}})Wang, Ji, Peng, Wu, Li, and
  Liu}]{wang-etal-2024-soft-knowledge}
Qunbo Wang, Ruyi Ji, Tianhao Peng, Wenjun Wu, Zechao Li, and Jing Liu.
  2024{\natexlab{d}}.
\newblock \href {https://doi.org/10.18653/v1/2024.acl-long.332} {Soft knowledge
  prompt: Help external knowledge become a better teacher to instruct {LLM} in
  knowledge-based {VQA}}.
\newblock In \emph{Proceedings of the 62nd Annual Meeting of the Association
  for Computational Linguistics (Volume 1: Long Papers)}, pages 6132--6143,
  Bangkok, Thailand. Association for Computational Linguistics.

\bibitem[{Wen et~al.(2024)Wen, Hovy, and Hauptmann}]{wen-etal-2024-transitive}
Haoyang Wen, Eduard Hovy, and Alexander Hauptmann. 2024.
\newblock \href {https://doi.org/10.18653/v1/2024.acl-long.80} {Transitive
  consistency constrained learning for entity-to-entity stance detection}.
\newblock In \emph{Proceedings of the 62nd Annual Meeting of the Association
  for Computational Linguistics (Volume 1: Long Papers)}, pages 1467--1480,
  Bangkok, Thailand. Association for Computational Linguistics.

\bibitem[{Yang et~al.(2023)Yang, Xiao, Wang, Zhang, Bian, Yin, Lv, Pan, Wang,
  Yan et~al.}]{yang2023baichuan}
Aiyuan Yang, Bin Xiao, Bingning Wang, Borong Zhang, Ce~Bian, Chao Yin, Chenxu
  Lv, Da~Pan, Dian Wang, Dong Yan, et~al. 2023.
\newblock Baichuan 2: Open large-scale language models.
\newblock \emph{arXiv preprint arXiv:2309.10305}.

\bibitem[{Yang et~al.(2024)Yang, Yang, Hui, Zheng, Yu, Zhou, Li, Li, Liu, Huang
  et~al.}]{yang2024qwen2}
An~Yang, Baosong Yang, Binyuan Hui, Bo~Zheng, Bowen Yu, Chang Zhou, Chengpeng
  Li, Chengyuan Li, Dayiheng Liu, Fei Huang, et~al. 2024.
\newblock Qwen2 technical report.
\newblock \emph{arXiv preprint arXiv:2407.10671}.

\bibitem[{Yao et~al.(2012)Yao, Bai, Liu, Ma, and Tu}]{yao2012detecting}
Cong Yao, Xiang Bai, Wenyu Liu, Yi~Ma, and Zhuowen Tu. 2012.
\newblock Detecting texts of arbitrary orientations in natural images.
\newblock In \emph{2012 IEEE conference on computer vision and pattern
  recognition}, pages 1083--1090. IEEE.

\bibitem[{Yi et~al.(2025)Yi, He, Wang, Song, Qian, Yuan, Zhang, Sun, Li, Lu
  et~al.}]{yi2025score}
Qiang Yi, Yangfan He, Jianhui Wang, Xinyuan Song, Shiyao Qian, Xinhang Yuan,
  Miao Zhang, Li~Sun, Keqin Li, Kuan Lu, et~al. 2025.
\newblock Score: Story coherence and retrieval enhancement for ai narratives.
\newblock \emph{arXiv preprint arXiv:2503.23512}.

\bibitem[{Young et~al.(2024)Young, Chen, Li, Huang, Zhang, Zhang, Li, Zhu,
  Chen, Chang et~al.}]{young2024yi}
Alex Young, Bei Chen, Chao Li, Chengen Huang, Ge~Zhang, Guanwei Zhang, Heng Li,
  Jiangcheng Zhu, Jianqun Chen, Jing Chang, et~al. 2024.
\newblock Yi: Open foundation models by 01. ai.
\newblock \emph{arXiv preprint arXiv:2403.04652}.

\bibitem[{Yu et~al.(2025)Yu, Lu, Wang, Li, Wang, Huang, and Li}]{yu2025eve}
Haiyang Yu, Jinghui Lu, Yanjie Wang, Yang Li, Han Wang, Can Huang, and Bin Li.
  2025.
\newblock Eve: Towards end-to-end video subtitle extraction with
  vision-language models.
\newblock \emph{arXiv preprint arXiv:2503.04058}.

\bibitem[{Yu et~al.(2016)Yu, Poirson, Yang, Berg, and Berg}]{yu2016modeling}
Licheng Yu, Patrick Poirson, Shan Yang, Alexander~C Berg, and Tamara~L Berg.
  2016.
\newblock Modeling context in referring expressions.
\newblock In \emph{Computer Vision--ECCV 2016: 14th European Conference,
  Amsterdam, The Netherlands, October 11-14, 2016, Proceedings, Part II 14},
  pages 69--85. Springer.

\bibitem[{Yu et~al.(2024)Yu, Gao, Yao, Wang, Ye, Wang, Xie, Zhang, and
  Zhang}]{yu-etal-2024-kieval}
Zhuohao Yu, Chang Gao, Wenjin Yao, Yidong Wang, Wei Ye, Jindong Wang, Xing Xie,
  Yue Zhang, and Shikun Zhang. 2024.
\newblock \href {https://doi.org/10.18653/v1/2024.acl-long.325} {{KIE}val: A
  knowledge-grounded interactive evaluation framework for large language
  models}.
\newblock In \emph{Proceedings of the 62nd Annual Meeting of the Association
  for Computational Linguistics (Volume 1: Long Papers)}, pages 5967--5985,
  Bangkok, Thailand. Association for Computational Linguistics.

\bibitem[{Yuliang et~al.(2017)Yuliang, Lianwen, Shuaitao, and
  Sheng}]{yuliang2017detecting}
Liu Yuliang, Jin Lianwen, Zhang Shuaitao, and Zhang Sheng. 2017.
\newblock Detecting curve text in the wild: New dataset and new solution.
\newblock \emph{arXiv preprint arXiv:1712.02170}.

\bibitem[{Zhao et~al.(2024)Zhao, Long, Liu, Kamoi, Nan, Chen, Liu, Tang, Zhang,
  and Cohan}]{zhao-etal-2024-docmath}
Yilun Zhao, Yitao Long, Hongjun Liu, Ryo Kamoi, Linyong Nan, Lyuhao Chen, Yixin
  Liu, Xiangru Tang, Rui Zhang, and Arman Cohan. 2024.
\newblock \href {https://doi.org/10.18653/v1/2024.acl-long.852}
  {{D}oc{M}ath-eval: Evaluating math reasoning capabilities of {LLM}s in
  understanding long and specialized documents}.
\newblock In \emph{Proceedings of the 62nd Annual Meeting of the Association
  for Computational Linguistics (Volume 1: Long Papers)}, pages 16103--16120,
  Bangkok, Thailand. Association for Computational Linguistics.

\bibitem[{Zhou et~al.(2024)Zhou, Li, Hong, Zhang, Wang, and
  Huai}]{zhou-etal-2024-copyne}
Shilin Zhou, Zhenghua Li, Yu~Hong, Min Zhang, Zhefeng Wang, and Baoxing Huai.
  2024.
\newblock \href {https://doi.org/10.18653/v1/2024.acl-long.147} {{C}opy{NE}:
  Better contextual {ASR} by copying named entities}.
\newblock In \emph{Proceedings of the 62nd Annual Meeting of the Association
  for Computational Linguistics (Volume 1: Long Papers)}, pages 2675--2686,
  Bangkok, Thailand. Association for Computational Linguistics.

\end{thebibliography}
\newpage
\twocolumn
\appendix

\section{Implementation}
\label{appendix:imple}

The proposed loss function is incorporated into the model's training objective through linear combination with a weighting coefficient $\lambda=0.3$. The hyperparameters governing the loss computation are maintained at $\alpha=\beta=\sigma=0.2$ throughout all experiments, unless otherwise specified. All tasks are trained with a learning rate of $10^{-5}$ for fine-tuning. Our experiments are conducted based on a widely used open source training repository\footnote{https://github.com/hiyouga/LLaMA-Factory}.

\section{Dataset}

\begin{figure*}[ht]
  \centering
  \includegraphics[width=0.85\textwidth]{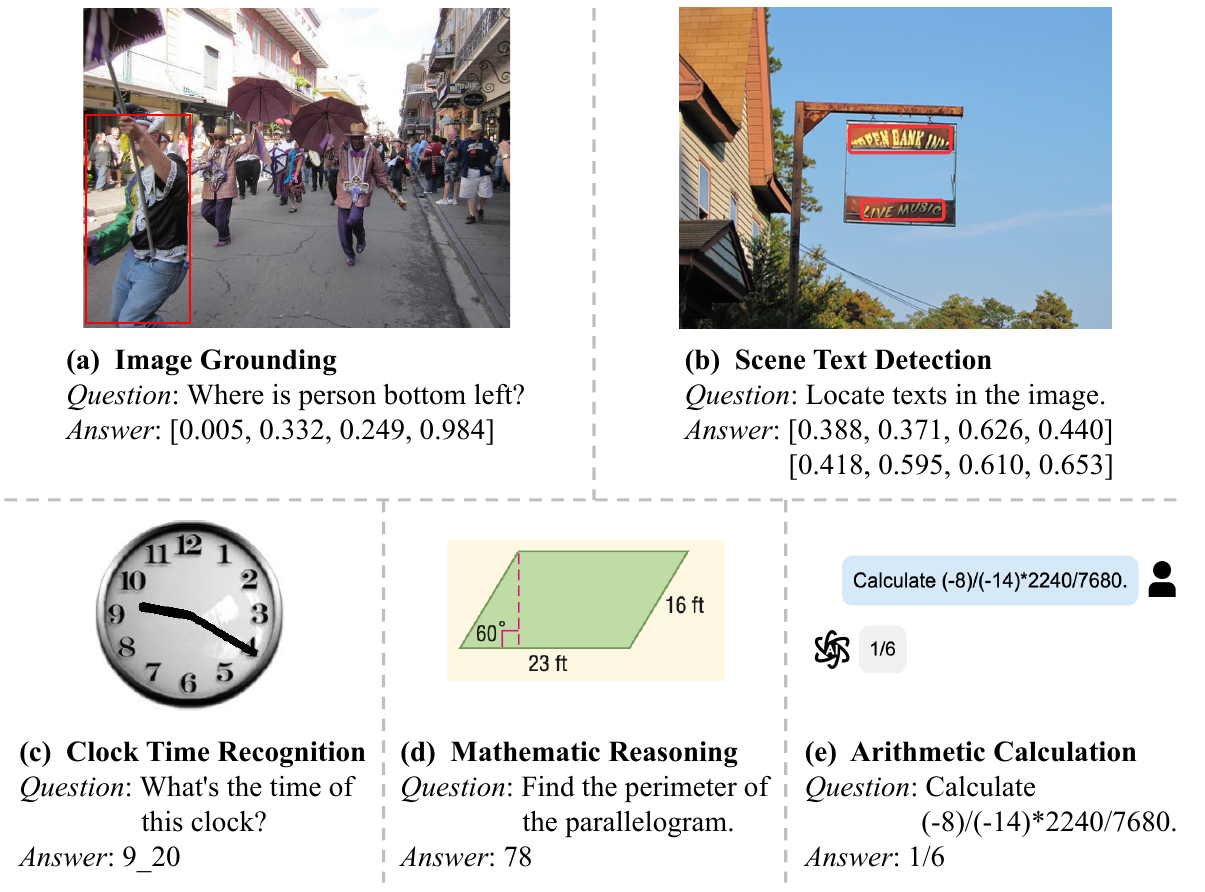}
    \caption{The illustrations for each task.}
    \label{fig:tasks}
\end{figure*}

This section provides a detailed description of each task along with the corresponding evaluation metrics. The illustrations for each task are presented in Figure~\ref{fig:tasks}.

       
       
       
       

\label{appendix:data}

\noindent \textbf{Image Grounding.} Grounding task aims to output the bounding box of the corresponding object given a description. We compare on the referring expression comprehension (REC) task on RefCOCO~\cite{lin2014microsoft}, RefCOCO+~\cite{yu2016modeling} and RefCOCOg~\cite{mao2016generation} datasets. The Average Accuracy at $\operatorname{IoU} \geq 0.5$ (Acc@0.5) is used as the evaluation metric.

\noindent \textbf{Scene Text Detection.} The scene text detection task focuses on detecting text in natural images. We selected several commonly used datasets: TD500~\cite{yao2012detecting}, ICDAR2015~\cite{karatzas2015icdar}, CTW1500~\cite{yuliang2017detecting} and Total-Text~\cite{ch2017total} for scene text detection tasks. We utilize the identical metric employed in the image grounding task.

\noindent \textbf{Clock Time Recognition.} The perception of clock aims to recognize the specific time by images of clocks. We compare the performance of accuracy and time gap on a widely-used TIME~\cite{time_dataset_kaggle} dataset. The output are formatted as the label of ``2\_55'', as shown in Figure~\ref{fig:goodcases}. We use overall accuracy as an metric, and additionally count the time gap between the prediction and the ground truth for further evaluation. For example, the time gap between prediction ``4\_35'' and ground truth ``6\_20'' is 1.75 hours.

\noindent \textbf{Mathematical Reasoning.} Completing the mathematical reasoning tasks requires models to understand the context and the image of the mathematical field. We select the MathVista~\cite{lu2023mathvista} and MathVision~\cite{wang2024measuring} datasets to evaluate models. We utilize exact matching accuracy to evaluate math reasoning task.

\noindent \textbf{Arithmetic Calculations.} Calculation task involves training LLMs to perform numerical operations accurately. In this task, the ``arithmetic\_mix'' subset from the widely-used mathematics dataset~\cite{saxton2019analysing} is used for training and evaluation, which contains 2M training and 10k test items. In this task, exact matching accuracy is applied as the evaluation metric. 


\section{Gumbel Softmax}\label{appendix:gumbel_softmax}

The Gumbel softmax, also known as Concrete Distribution, is a continuous differentiable approximation to categorical sampling. It replaces the non-differentiable $\operatorname{argmax}$ operation with a softmax function and Gumbel noise. Given logits $\pi_i$, the Gumbel softmax sample $y_i$ is computed as: 
$$
y_i = \operatorname{softmax}\left((log(\pi_i) + g_i)/\tau\right),
$$
where $g_i$ is the Gumbel noise, which is i.i.d. samples drawn from the $\operatorname{Gumbel}(0,1)$ distribution, and $\tau$ is the temperature parameter. 

The Gumbel noise term $g_i$ introduces stochasticity into the sampling process, enabling exploration of the probability space while maintaining differentiability. Moreover, using Gumbel noise also works like regularization, which helps provide gradient information near the decision boundary, to improve generalization ability.
The temperature parameter $\tau$ controls the sharpness of the distribution: as $\tau$ approaches 0, the samples become more discrete and closer to one-hot vectors, while higher temperatures make the distribution more uniform. In our implementation, we use $\tau=0.1$ to ensure that the results are consistent with the original $\operatorname{argmax}$ results.

Gumbel softmax is differentiable as it replaces the discrete $\operatorname{argmax}$ with a continuous softmax function, allowing gradients to flow through the sampling process during backpropagation. 
Thus, Gumbel softmax is widely used in scenarios requiring discrete latent variables in neural networks, such as in VAEs\cite{jang2016categorical} or reinforcement learning\cite{huijben2022review,wan2020fbnetv2}.

\section{Preliminaries}
\label{sec:preliminaries}

This section first briefly introduces the autoregressive decoding process based on cross-entropy in Section~\ref{sec:prelim_2.1}, and then compares and analyzes Earth Mover's Distance (EMD) in Section~\ref{sec:prelim_2.2}.

\subsection{Autoregressive Prediction with Cross Entropy}
\label{sec:prelim_2.1}

Autoregressive models operate through sequential decoding, generating tokens one at a time conditioned on previously generated tokens. For each position, the model outputs a probability distribution across the vocabulary, employing the Softmax function to select the most probable token during training.

In the context of language modeling tasks, cross-entropy loss serves as the fundamental training objective for autoregressive models. This loss function quantifies the divergence between the predicted probability distribution and the ground truth distribution:
\begin{equation}
\mathcal{L}=-\sum_i p_i \log \left(q_i\right),
\label{eqa:cross_entropy}
\end{equation}
Åwhere $p_i$ represents the one-hot encoded ground truth distribution, and $q_i$ denotes the model's predicted probability.

While cross-entropy loss effectively minimizes distributional differences between predictions and labels during training, it exhibits a fundamental limitation in autoregressive decoding: the function treats each class independently, disregarding the inherent relationships between different classes. This limitation becomes particularly problematic when modeling numerical sequences where ordinal relationships between values carry semantic significance\cite{hou2016squared}, as shown in Figure~\ref{fig:emd_example}.

\subsection{Earth Mover's Distance}
\label{sec:prelim_2.2}

To introduce a distance term when calculating the above-mentioned distribution differences, one method is Earth Mover's Distance (EMD), also known as Wasserstein distance. It is an evaluation based on optimal transport theory, measuring the minimal cost of transforming one distribution into the other:
\begin{equation}
\operatorname{EMD}(P, Q)=\min _{\gamma \in \Gamma(P, Q)} \sum_{i=1}^n \sum_{j=1}^m \gamma_{i j} \cdot d\left(x_i, y_j\right),
\label{equ:emd_base}
\end{equation}
where $P=\left\{\left(p_i, x_i\right)\right\}$ and $Q=\left\{\left(q_i, y_i\right)\right\}$ are two discrete distributions, with $p_i$ and $q_j$ are the masses at the points $x_i$ and $y_j$, respectively. The transport plans, represented as $\Gamma(P, Q)$, are all possible ways to move the mass, and $\gamma_{i j}$ represents the amount of mass that is transported from $p_i$ to $q_j$. The distance matrix $d\left(x_i, y_j\right)$ indicates the cost of transporting masses between points $x_i$ and $y_j$. A widely-used distance matrix $d$ is Euclidean distance.


Since the distance between labels is explicitly considered, predicted values closer to the label are associated with smaller distance terms. Thus, the Earth Mover's Distance effectively incorporates distance-based weighting. As illustrated in Figure~\ref{fig:emd_example}, when the distribution is more concentrated around the label, the EMD loss becomes smaller, thereby reflecting the differences between distributions.

\subsection{Predicting Digits with EMD}
\label{sec:method_3.1}

This section presents our approach to refining distance metrics for numerical representation at the digit level. In traditional autoregressive models, cross-entropy loss is typically employed to predict the probability distributions of individual tokens. However, this method treats each numerical digit as an independent entity, disregarding the continuous relationships between numbers. For example, when the target digit is $4$, a model prediction of $3$ should ideally be considered closer to accurate than a prediction of $9$, as it represents a smaller numerical deviation. To address this limitation, we propose incorporating a distance metric that captures these intrinsic numerical relationships more accurately.

\noindent \textbf{Computational Complexity.} As established in~\ref{sec:prelim_2.2}, Earth Mover's Distance (EMD) provides a robust measure for distributional distances, making it particularly well-suited for numerical prediction tasks. 
Prior research has applied EMD to align hidden representations within neural networks, often requiring the transport plan ($\gamma_{i j}$ in Equ.~\eqref{equ:emd_base}) to be approximated or recalculated dynamically during training. However, the computational demands of EMD present practical challenges, especially in large-scale deep learning applications. Solving the underlying optimization problem in Equ.~\eqref{equ:emd_base} has a computational complexity of $O\left((n \times m)^3\right)$, which can be prohibitive. Regularized EMD~\cite{cuturi2013sinkhorn} addresses this by employing the Sinkhorn-Knopp algorithm to iteratively refine the transport plan $\gamma_{i j}$ in Equ.~\eqref{equ:emd_base}, reducing complexity to $O\left(k \times n \times m\right)$, where each iteration involves an $O(n \times m)$ matrix operation. 

\noindent \textbf{Numerical Prediction Optimization with EMD.} When estimating the transport plan, the algorithm’s complexity is generally quadratic. However, when restricted to one-dimensional numerical distributions, where the prediction and target values are aligned in position ($i=j$), the transport plan can be simplified to an identity matrix. Thus, Earth Mover's Distance emerges as a highly suitable metric for capturing digit-level numerical distance, formulated as:
\begin{equation}
\operatorname{EMD}(P, Q)= \sum_{i} \left|x_i - y_i\right| \cdot \left|i - \operatorname{argmax}\left(Q\right) \right|,
\label{equ:emd_digit}
\end{equation}
where the distance matrix $d\left(x_i, y_j\right) = \left|i - \operatorname{argmax}\left(Q\right) \right|$ refers to the index distance of each digit to the label. 
Given that the predicted probability distribution $P$ is obtained through the softmax transformation, and the ground truth label $Q$ is represented as a one-hot vector, the gradient of EMD with respect to component $x_i$ can be expressed as:
\begin{equation}
\frac{\partial \operatorname{EMD}}{\partial x_i} = \left\{ \left|k-1\right|, \left|k-2\right|, ..., \left|k-n\right| \right\}.
\end{equation}
where $k=\operatorname{argmax}\left(Q\right)$ denotes the index of the label element in the one-hot vector. This gradient exhibits an inverse relationship with the proximity between the predicted distribution and the ground truth: as the prediction approaches the true label, the magnitude of the gradient diminishes. 
This characteristic is particularly advantageous for numerical prediction tasks, as it inherently accounts for the ordinal relationships between numerical classes, and addresses the fundamental limitation of the conventional cross-entropy.



\begin{table}[t]
  \small
  \centering
  \scriptsize
  \resizebox{\columnwidth}{!}{%
  \begin{tabular}{c|ccccc}
  \hline
  $\beta$ $\backslash$ $\alpha$ & 0 & 0.1 & 0.2 & 0.3 & 0.4 \\
  \hline
  0 & 0.780 & 0.783 & 0.789 & 0.789 & 0.771 \\
  0.1 & 0.777 & \textbf{0.790} & \textbf{0.799} & 0.789 & 0.778 \\
  0.2 & 0.782 & \textbf{0.793} & \textbf{0.795} & 0.785 & 0.784 \\
  0.3 & 0.788 & 0.783 & 0.786 & 0.785 & 0.781 \\
  0.4 & 0.774 & 0.782 & 0.778 & 0.781 & 0.784 \\
  \hline
  \end{tabular}
  }
  \caption{Ablation studies of hyper-parameters $\alpha$ and $\beta$}
  \label{tab:ablation_alpha_beta}
\end{table}
  
\begin{table}[t]
  \small
  \centering
  \scriptsize
  \resizebox{\columnwidth}{!}{%
  \begin{tabular}{c|cccccccc}
  \hline
    & 0.0 & 0.1 & 0.2 & 0.3 & 0.4 & 0.5 & 0.6 & 0.7 \\
  \hline
  $\sigma$ & 0.784 & 0.785 & \textbf{0.789} & \textbf{0.788} & 0.783 & 0.780 & 0.770 & 0.763 \\
  $\lambda$ & 0.741 & 0.777 & \textbf{0.798} & \textbf{0.795} & 0.788 & 0.786 & 0.787 & 0.777 \\
  \hline
  \end{tabular}
  }
  \caption{Ablation studies of the hyper-parameters $\sigma$ (exp-weighting) and $\lambda$ (coefficient)}
  \label{tab:ablation_sigma_lambda}
\end{table}

\begin{table}[t]
  \small
  \centering
  \scriptsize
  \resizebox{\columnwidth}{!}{%
  \begin{tabular}{llccc}
  \hline
  Model & Params & CE & EMD & \textbf{NTIL} \\ \hline
  Qwen2.5 & 2b & 0.509 & 0.517 & \textbf{0.523} \\
  MiniCPM3 & 4b & 0.672 & 0.676 & \textbf{0.704} \\
  Yi & 6b & 0.375 & 0.383 & \textbf{0.400} \\
  LLaMA3 & 8b & 0.638 & 0.643 & \textbf{0.646} \\
  \hline
  \end{tabular}
  }
  \caption{Results of the GSM8k mathematical reasoning task.}
  \label{tab:gsm8k}
\end{table}

\section{Ablations}\label{appendix:aba}

We have supplemented our work with comprehensive experiments on the clock time recognition task 
using Qwen2-VL-7B, as shown in Table~\ref{tab:ablation_alpha_beta} and Table~\ref{tab:ablation_sigma_lambda}. 
We have bolded some of the outperforming results.
The results demonstrate that NTIL adapts well to different hyperparameters.

We have also supplemented NTIL with the GSM8k math reasoning dataset.
NTIL brings consistent performance improvement on mixed text-numeric tasks, 
which shows the effectiveness of our method.

\section{Qualitative Examples}\label{appendix:quali}

\begin{figure*}[ht]
    \centering
    \subfigure[Example in Image Grounding. Blue box is CE prediction, red box is EMD prediction, green box is NTIL prediction. Black box is ground truth.]{        \vtop{\vskip10pt\hbox{\includegraphics[width=0.95\textwidth]{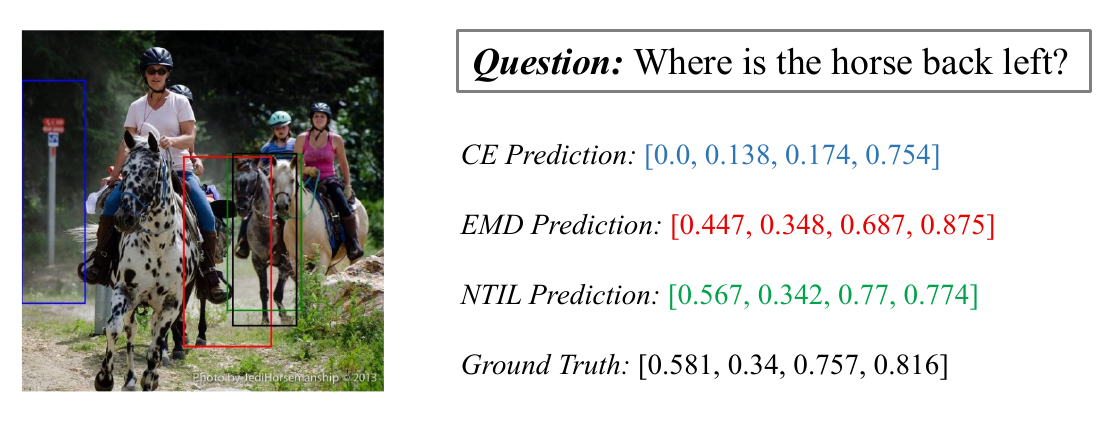}}}\label{subfig:image_ground}}
    \vspace{20pt}
    \subfigure[Example in clock time recognition. ]{        \vtop{\vskip10pt\hbox{\includegraphics[width=0.95\textwidth]{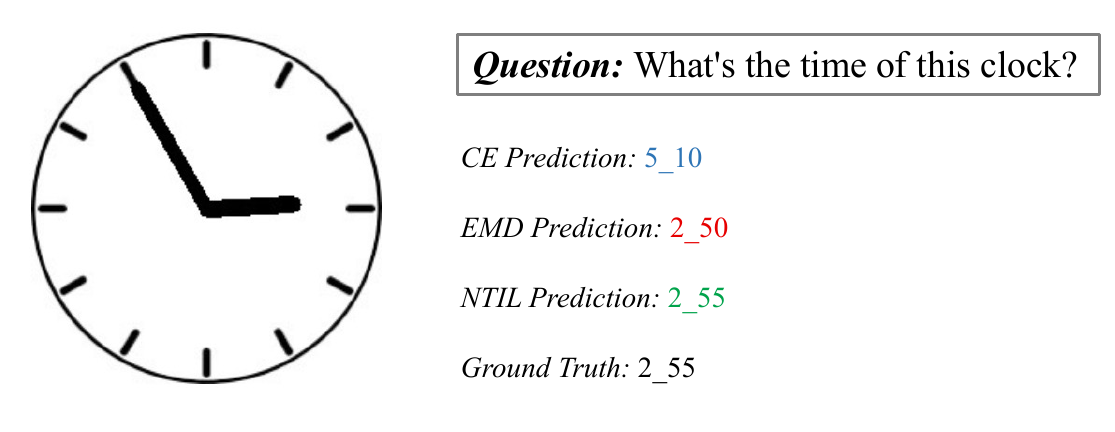}}}
       
    }
    \caption{Comparisons between CE, EMD and NTIL.}
    \label{fig:goodcases}
\end{figure*}

Visualizations of the outputs of different losses are shown in Figure~\ref{fig:goodcases}, and the examples are taken from experimental results using LLaVA-1.5. For image grounding task (Figure~\ref{subfig:image_ground}), the task was to predict the location of ``horse back left'' in an image. The CE loss (blue box) performed poorly, with predictions far from the ground truth. EMD (red box) showed an improvement, capturing spatial features better, while NTIL (green box) provided the most accurate predictions, closely matching the ground truth (black box). Overall, NTIL outperformed both CE and EMD, demonstrating its effectiveness in this task.

Figure~\ref{fig:goodcases}(b) presents a qualitative comparision for clock time recognition task. In this case, NTIL provides the most accurate prediction of the clock time, correctly identifying 2:55, which matches the ground truth. EMD performs better than CE, predicting 2:50, but it is still slightly off. CE, however, predicts 5:10, a significant deviation. Overall, NTIL outperforms both EMD and CE in predicting the clock time accurately.

\end{document}